\documentclass{article}
\pdfoutput=1

    \PassOptionsToPackage{numbers, compress}{natbib}
\usepackage[preprint]{neurips_2022}

\usepackage[utf8]{inputenc} % allow utf-8 input
\usepackage[T1]{fontenc}    % use 8-bit T1 fonts
\usepackage{hyperref}       % hyperlinks
\usepackage{url}            % simple URL typesetting
\usepackage{booktabs}       % professional-quality tables
\usepackage{amsfonts}       % blackboard math symbols
\usepackage{nicefrac}       % compact symbols for 1/2, etc.
\usepackage{microtype}      % microtypography
\usepackage{xcolor}         % colors
\usepackage{amsmath}
\usepackage{amsthm}
\usepackage{graphicx}
\usepackage{subcaption}
\usepackage{wrapfig}
\usepackage{dsfont}
\usepackage{multirow}
\usepackage{cancel}
\usepackage{pgfplots}
\usepackage{threeparttable}

\usepackage{tikz}
\usetikzlibrary{positioning,fit,calc}
\usepgfplotslibrary{groupplots}
\pgfplotsset{width=7.5cm,compat=1.15, tick label style={font=\small}}

\usepackage{xspace}

\makeatletter
\DeclareRobustCommand\onedot{\futurelet\@let@token\@onedot}
\def\@onedot{\ifx\@let@token.\else.\null\fi\xspace}

\def\eg{\emph{e.g\onedot}} 
\def\ie{\emph{i.e\onedot}} 
\def\cf{\emph{c.f}\onedot}

\makeatother

\usepackage{amsmath,amsfonts,bm}

\def\eqref#1{equation~\ref{#1}}
\def\1{\bm{1}}

\DeclareMathAlphabet{\mathsfit}{\encodingdefault}{\sfdefault}{m}{sl}
\SetMathAlphabet{\mathsfit}{bold}{\encodingdefault}{\sfdefault}{bx}{n}

\makeatletter
\newcommand*\readcoords[1]{\input{#1} }
\makeatother

\definecolor{cornellred}{RGB}{179,27,27}
\definecolor{cornellblue}{RGB}{0,104,172}
\definecolor{cornellgreen}{RGB}{110,180,63}
\definecolor{cornellgrey}{RGB}{85,86,90}
\definecolor{mypurple}{RGB}{116,79,198}

\title{Out-of-Distribution Detection with Class Ratio Estimation}

\author{%
   Mingtian Zhang\textsuperscript{14}\footnotemark[1]\thanks{Equal Contribution, the work was done during an internship in Huawei Noah's Ark Lab.}  \And Andi Zhang\textsuperscript{24}\footnotemark[1]\And Tim Z. Xiao\textsuperscript{3} \And Yitong Sun 
   \textsuperscript{4}\And Steven McDonagh\textsuperscript{4} 
   \Aff
  \textsuperscript{1}Centre for Artificial Intelligence, University College London\\
    \textsuperscript{2}Department of Computer Science and Technology, University of Cambridge\\
    \textsuperscript{3}University of Tübingen \& IMPRS-IS\quad 
    \textsuperscript{4}Huawei Noah’s Ark Lab,\\
\texttt{m.zhang@cs.ucl.ac.uk}\quad \texttt{az381@cam.ac.uk}\quad
\texttt{zhenzhong.xiao@uni-tuebingen.de} \\
\texttt{\{sunyitong,steven.mcdonagh\}@huawei.com}
}

\begin{document}

\maketitle

\begin{abstract}
Density-based Out-of-distribution (OOD) detection has recently been shown %to be 
unreliable for the task of detecting OOD images.  
Various density ratio based approaches 
achieve good empirical performance, however 
methods typically 
lack a principled probabilistic modelling explanation. In this work, we propose to unify density ratio based methods under a novel framework that builds energy-based models and employs differing base distributions. Under our framework, the density ratio can %also 
be viewed as the unnormalized density of an implicit semantic distribution.
Further, we propose to directly estimate the density ratio of a data sample through class ratio estimation. We report competitive results on OOD image problems in comparison with recent work that alternatively requires training of deep generative models for the task. Our approach enables a simple and yet effective path towards solving the OOD detection problem.
\end{abstract}

\section{Background}
\label{sec:background}

Machine learning methods often assume that training and testing data originate from the same distribution. However, in many real world applications, we usually have little control over the data source. Therefore, detecting Out-of-distribution (OOD) data is critical for safe and reliable machine learning applications. The importance of the problem has led to the design of many methods which aim to tackle OOD detection.  

Several works study the OOD detection problem in the scenario where both the in-distribution (ID) data and corresponding labels are presented. For example, \cite{hendrycks2016baseline} presents a practical baseline for OOD detection using a softmax confidence score. A selection of follow-up works then propose various routes to improve the performance of OOD detection \eg~using neural network calibration~\citep{guo2017calibration, kull2019beyond}, deep ensembles~\citep{lakshminarayanan2017simple}, the ODIN score~\citep{liang2017enhancing}, the Mahalanobis distance~\citep{lee2018simple}, energy based scores~\citep{liu2020energy} and Gram Matrices~\citep{sastry2019detecting}. However, different classes of labels require the classifier to encode different features. As a thought experiment, consider being given MNIST digits with binary labels that represent `\emph{if the 100th pixel is black or not}'. % then the 
Features learned by %from 
a resulting classifier will clearly not capture any semantic information %of the 
pertaining to %pertains to the 
digits. 
Therefore, approaches of this type strongly rely on the labels' quality, %which 
and we refer to this strategy as `label-dependent OOD detection'.

Alternatively, OOD detection can %also 
be performed without labels, where we hope the detector can distinguish between the ID and OOD data by using their `semantics'~\cite{ren2019likelihood, havtorn2021hierarchical}. Compared to %the 
label-dependent OOD, this type of OOD detection 
 is more consistent with human intuition and can potentially exhibit better generalisation and robustness.
One historically popular approach is to formalise the unsupervised OOD task as a density estimation problem \cite{bishop1994novelty}, which usually includes training a model on the ID data and using the evaluated density as the score for the OOD detection task. However, recent work \cite{nalisnick2018deep} shows that for popular deep generative models, OOD data may have a higher density (likelihood) value than ID data, which makes OOD detection fail.
Investigating further, various papers make proposals: typical sets~\cite{nalisnick2019detecting}, dataset complexity~\cite{serra2019input}, inductive bias~\cite{kirichenko2020normalizing}, likelihood domination \cite{schirrmeister2020understanding,zhang2020spread} towards explaining this phenomenon, yet the fundamental reasons underlying why deep generative models fail to detect OOD datasets remain not fully understood~\cite{kirichenko2020normalizing}. 

Recently, many density ratio methods have been proposed to resolve this challenge~\citep{ren2019likelihood, kirichenko2020normalizing, havtorn2021hierarchical, zhang2021out}, where the ratio %of 
between %the 
density values, which are evaluated by two models respectively, is used %to be 
as the score function for the OOD detection. Although good empirical results are achieved in practice, such approaches usually lack a principled modelling explanation. Additionally, we observe the proposed methods all require training exactly one or two generative models, which is computationally expensive and potentially redundant if the goal is to estimate the density ratios. Therefore, in this work, we %are aiming 
aim to understand the density ratio-based methods from a modelling perspective and show how to estimate the density without training generative models.

Explicitly, our main contributions are as follows:
\begin{itemize}
    \item We unify the existing density ratio based OOD detection methods as building energy based models on the in-distribution data.  
    \item We propose to use class ratio estimation to evaluate density ratios with a binary classifier. This significantly simplifies the procedure of learning an OOD detector yet retains competitive performance. 
    \item Through the consideration of a classical class ratio estimation problem, we propose a corresponding solution to boost performance for practical OOD detection instances.
\end{itemize}
In the following section, we first briefly introduce OOD detection. 

%\section{Introduction of OOD Detection}
\section{OOD Detection}
\label{sec:ood_detection}
Given an in-distribution dataset $\mathcal{D}_{\text{in}}=\{x_1,\ldots, x_N\}$, we would like to learn an OOD detector which can be formalised as an indicator 
function that maps a data $x$ to $\{0,1\}$: %, %which can be formalized as
 %such that %follows 
\begin{equation*}
    D_\epsilon(x) = 
    \begin{cases}
        0 & \quad s(x) < \epsilon~;\\
        1 & \quad s(x) \ge \epsilon~,
    \end{cases}
\end{equation*}
where $\epsilon$ is a hyper-parameter which represents the confidence threshold and $s(x)$ is a score function representing whether or not a data $x$ is likely to be an in-distribution  sample. %or not. %sample from the in-distribution or not. 
In practice, the threshold $\epsilon$ can be determined \eg~using the validation dataset. For evaluation, the Area Under the Receiver Operating Characteristics (AUROC) is calculated using the ID and OOD test datasets~\cite{hendrycks2016baseline}, %which 
automatically %takes the 
incorporating the consideration of different choices of $\epsilon$ value. Higher AUROC indicates that the detector has a better ability to discriminate between ID and OOD data.

\subsection{Density-based OOD Detection}
\label{sec:ood_detection:density}
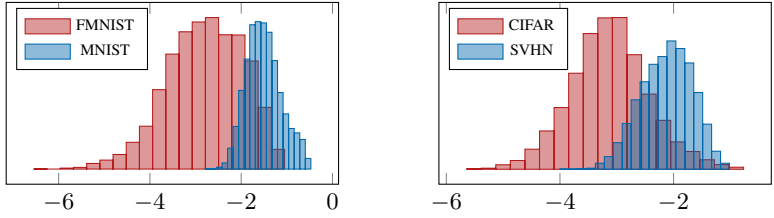
\begin{figure}[t]
  \begin{minipage}[t]{1.\linewidth}
  \centering
  \begin{subfigure}{.4\textwidth}
    \centering
    \begin{tikzpicture}
        \begin{axis}[
            width=6cm, height=4cm, area style,  legend pos=north west, font=\small, ymajorticks=false]
        \addplot+[ybar interval,mark=no, fill opacity=0.45, color=cornellred] plot coordinates { 
(-6.542270183563232, 1)
(-6.253206494450569, 0)
(-5.9641428053379055, 8)
(-5.675079116225243, 14)
(-5.3860154271125795, 47)
(-5.096951737999916, 75)
(-4.807888048887253, 122)
(-4.518824359774589, 224)
(-4.229760670661927, 359)
(-3.9406969815492627, 665)
(-3.6516332924365997, 963)
(-3.3625696033239363, 1150)
(-3.073505914211273, 1232)
(-2.78444222509861, 1269)
(-2.495378535985947, 1125)
(-2.2063148468732834, 1123)
(-1.91725115776062, 905)
(-1.6281874686479565, 517)
(-1.339123779535293, 165)
(-1.0500600904226305, 36)
        };
        \addlegendentry{\tiny FMNIST}
        \addplot+[ybar interval,mark=no, fill opacity=0.45, color=cornellblue] plot coordinates { 
(-2.7865819931030273, 2)
(-2.6648542165756224, 2)
(-2.543126440048218, 15)
(-2.421398663520813, 51)
(-2.2996708869934084, 175)
(-2.1779431104660034, 357)
(-2.0562153339385985, 659)
(-1.934487557411194, 921)
(-1.812759780883789, 1209)
(-1.6910320043563842, 1235)
(-1.5693042278289795, 1224)
(-1.4475764513015748, 1146)
(-1.32584867477417, 909)
(-1.2041208982467653, 641)
(-1.0823931217193603, 462)
(-0.9606653451919556, 340)
(-0.8389375686645508, 303)
(-0.7172097921371461, 248)
(-0.5954820156097411, 89)
(-0.4737542390823366, 12)
        };
        \addlegendentry{\tiny MNIST}
        \end{axis}
    \end{tikzpicture}
    % \caption{Fashion vs MNIST}
  \end{subfigure}%
  \begin{subfigure}{.4\textwidth}
    \centering
    \begin{tikzpicture}
        \begin{axis}[
            width=6cm, height=4cm, area style,  legend pos=north west, font=\small, ymajorticks=false]
        \addplot+[ybar interval,mark=no, fill opacity=0.45, color=cornellred] plot coordinates { 
(-5.642870933354154, 5)
(-5.386297750018464, 11)
(-5.129724566682773, 46)
(-4.873151383347083, 90)
(-4.616578200011393, 207)
(-4.360005016675702, 395)
(-4.103431833340012, 636)
(-3.8468586500043216, 990)
(-3.590285466668631, 1399)
(-3.3337122833329405, 1561)
(-3.0771390999972503, 1564)
(-2.8205659166615598, 1181)
(-2.563992733325869, 775)
(-2.307419549990179, 504)
(-2.0508463666544885, 282)
(-1.7942731833187984, 183)
(-1.5376999999831078, 98)
(-1.2811268166474177, 48)
(-1.0245536333117267, 21)
(-0.7679804499760365, 4)
        };
        \addlegendentry{\tiny CIFAR}
        \addplot+[ybar interval,mark=no, fill opacity=0.45, color=cornellblue] plot coordinates { 
(-4.005874938344816, 2)
(-3.8485297079659206, 2)
(-3.6911844775870253, 7)
(-3.53383924720813, 11)
(-3.376494016829235, 62)
(-3.21914878645034, 130)
(-3.0618035560714447, 274)
(-2.9044583256925494, 472)
(-2.7471130953136544, 716)
(-2.5897678649347595, 902)
(-2.432422634555864, 1082)
(-2.275077404176969, 1162)
(-2.117732173798074, 1323)
(-1.960386943419179, 1245)
(-1.8030417130402836, 1096)
(-1.6456964826613882, 791)
(-1.4883512522824933, 465)
(-1.3310060219035984, 186)
(-1.173660791524703, 62)
(-1.0163155611458077, 10)
        };
        \addlegendentry{\tiny SVHN}
        \end{axis}
    \end{tikzpicture}
    % \caption{CIFAR10 vs SVHN} 
  \end{subfigure}
  \caption{The left plot shows a PixelCNN model that is trained on FashionMNIST and tested on FashionMNIST (ID) and MNIST (OOD); the right plot show a PixelCNN model that is trained on CIFAR10 and tested on CIFAR10 (ID) and SVHN (OOD). The $x$-axis indicates the log-likelihood normalised by the data dimension and $y$-axis represents the data counts. We can observe that OOD datasets consistently obtain higher test likelihood than ID datasets. Plots are derived from~\cite{zhang2021out}. \label{fig:ood:example}}
\end{minipage}\hfill
\end{figure}

For the unsupervised OOD detection problem,
a natural strategy is to learn a model $p_\theta(x)$ to fit the in-distribution dataset $\mathcal{D}_{\text{in}}$. The parameters $\theta$ can be learned by  Maximum Likelihood Estimation
\begin{align}
    \theta^*=\arg\max_\theta \frac{1}{N}\sum_{n=1}^N p_\theta(x_n).
\end{align}
For a given test data $x'$, the density evaluation under the learned model $p_{\theta^*}(x')$ can be used as the score function for the OOD detection $s(x')\equiv p_{\theta^*}(x')$. In this case, lower density indicates that test data is more likely to be OOD~\citep{bishop1994novelty}. Popular choices for the model $p_{\theta}$ are deep generative models such as Flow models \citep{kingma2018glow, kirichenko2020normalizing}, latent variable models~\citep{kingma2013auto} or auto-regressive models \citep{salimans2017pixelcnn++}. However, recent work~\citep{nalisnick2018deep} shows the surprising result that deep generative models may assign \emph{higher} density values to OOD data, that contain differing semantics, \cf~the ID data that was used for maximum likelihood training. Figure \ref{fig:ood:example} shows an example of this effect where PixelCNN models trained on Fashion MNIST, CIFAR10 induce higher test likelihoods %that are 
when evaluated on MNIST, SVHN respectively. Recently, many density ratio based methods \citep{ren2019likelihood, kirichenko2020normalizing, havtorn2021hierarchical, zhang2021out} are proposed, and achieve empirical success, where the score function is usually defined as the density ratio between two generative models $p,q$ with different model structures. In the next section, we propose an energy-based model framework which %allows to provide 
enables a unified view of %the 
density ratio methods.

\section{Unifying Density Ratio Methods with Energy-based Models}
\label{sec:ebm}

Recent work~\cite{zhang2021out} proposes to model the in-distribution data using a product of local and non-local models and show that the non-local model can be considered a model of the data semantics. Here, we generalise this idea and define a general energy-based model for the in-distribution data, which in turn allows us to unify other density ratio-based OOD detection methods, such that they may be viewed as implicitly building semantic models on the in-distribution dataset.

We propose to model the in-distribution $p_{\text{in}}$ with an energy-based model that can be defined as %follows
\begin{align}
    p_{\text{in}}(x) = \frac{p_{\text{base}}(x) s(x)}{Z_{\text{in}}}, \quad \text{with} \quad Z_{in}=\int p_{\text{base}}(x)s(x)dx, \label{eq:in:distribution}
\end{align}
where $p_{\text{base}}$ is the base distribution and $s(x)$ is a positive function that gives high score for the image $x$ whose semantics belongs to the in-distribution $p_{\text{in}}$, such that the score function may be thought of in this case as a %which we referred to the 
`semantic score'.
A semantic distribution $p_s(x)$ can then be further defined as the normalised score function 
\begin{align}
    p_{s}(x) = \frac{s(x)}{Z_s}, \quad \text{with} \quad Z_s=\int s(x)dx. \label{eq:semantic:model}
\end{align}

\begin{wrapfigure}{R}{0.4\textwidth}
\center
\vspace{-0.6cm}
\begin{tikzpicture}[scale=1.0]
\begin{scope}[thick, cornellred]
\node at (3.9, -3.5) {$p_{\text{base}}(x)$};
\draw (1.7352941176470589, -0.46472172941176476) .. controls (2.5069289411764704, -0.4698823176470588) and (2.706279294117647, -1.2877097647058824) .. (2.8411764705882354, -1.6294276470588236);
\draw (2.8411764705882354, -1.6294276470588236) .. controls (3.1132082352941177, -2.2959378823529413) and (3.4529411764705884, -2.270604117647059) .. (3.458823529411765, -2.770604117647059);
\draw (3.458823529411765, -2.770604117647059) .. controls (3.466871176470588, -3.111527882352941) and (3.130025294117647, -3.698536235294118) .. (2.3529411764705883, -3.7117805882352943);
\draw (2.3529411764705883, -3.7117805882352943) .. controls (1.561421882352941, -3.7394600000000002) and (0.4904727529411765, -2.9754037647058826) .. (0.48823529411764705, -1.870604117647059);
\draw (0.48823529411764705, -1.870604117647059) .. controls (0.5004330588235294, -0.8163276470588235) and (1.341259294117647, -0.4619736117647059) .. (1.7352941176470589, -0.46472172941176476);
\end{scope}
\begin{scope}[thick, cornellblue]
\node at (1.7, -1.375) {$p(x)$};
\draw (1.4536450588235295, -0.8235294117647058) .. controls (1.6060144705882355, -0.8250729411764707) and (1.7564770588235292, -0.9262461176470588) .. (1.8828397647058823, -0.990556);
\draw (1.8828397647058823, -0.990556) .. controls (2.062590823529412, -1.0767088235294118) and (2.3021604705882353, -1.1814776470588235) .. (2.3014631764705884, -1.392265294117647);
\draw (2.3014631764705884, -1.392265294117647) .. controls (2.2986615294117647, -1.6266757647058823) and (2.0196441176470588, -1.698729176470588) .. (1.8828397647058823, -1.7559179999999999);
\draw (1.8828397647058823, -1.7559179999999999) .. controls (1.7460355294117647, -1.8131068235294119) and (1.5359714117647059, -1.8783116470588235) .. (1.407131294117647, -1.8764308235294118);
\draw (1.407131294117647, -1.8764308235294118) .. controls (1.1209019999999998, -1.875717411764706) and (1.017134705882353, -1.6889362352941177) .. (1.011764705882353, -1.5148923529411764);
\draw (1.011764705882353, -1.5148923529411764) .. controls (1.0119463529411765, -1.4321119999999998) and (1.053453294117647, -1.3490981176470587) .. (1.098449411764706, -1.2421529411764705);
\draw (1.098449411764706, -1.2421529411764705) .. controls (1.1641989411764706, -1.0962948235294117) and (1.2559863529411766, -0.8236844705882354) .. (1.4536450588235295, -0.8235294117647058);
\end{scope}
\begin{scope}[thick, cornellblue]
\node at (2.2, -2.6) {$q(x)$};
\draw (2.341763294117647, -1.9589796470588237) .. controls (2.6968315294117646, -1.9697523529411762) and (2.7692944705882354, -2.1916709411764708) .. (2.7874101176470587, -2.491153294117647);
\draw (2.7874101176470587, -2.491153294117647) .. controls (2.805525882352941, -2.7906356470588234) and (2.522920470588235, -3.2) .. (2.1787216470588233, -3.2);
\draw (2.1787216470588233, -3.2) .. controls (1.8345228235294118, -3.2) and (1.334528705882353, -3.0448724705882353) .. (1.323659294117647, -2.771244705882353);
\draw (1.323659294117647, -2.771244705882353) .. controls (1.3127898823529411, -2.4976169411764704) and (1.9866949411764705, -1.9482069411764706) .. (2.341763294117647, -1.9589796470588237);
\end{scope}
\end{tikzpicture}
\caption{A visualisation of the energy based model. The real lines indicate the \emph{high-density regions} of the distributions.\label{fig:pinpbase}}
\vspace{-0.5cm}
\end{wrapfigure}
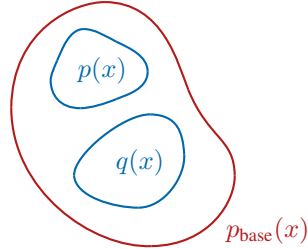

The density of the semantic distribution $p_s(x)$ can then be used to conduct semantic-level OOD detection. In practice, the estimation of the normalisation constant is not necessary for OOD detection tasks, since the score function is proportional to the likelihood ratio such that 
\begin{align}
    s(x)\propto\frac{p_{\text{in}}(x)}{p_{\text{base}}(x)},\label{eq:density:ratio}
\end{align}
so utilising the density ratio; \mbox{$p_{\text{in}}(x)/p_{\text{base}}(x)$} is equivalent to using the semantic distribution $p_s(x)$  density value, in the context of OOD detection.

This model formulation is partially inspired by recent work~\cite{arbel2020generalized,che2020your}, where an energy-based model is defined using a trained Generative Adversarial Network (GAN)~\cite{goodfellow2014generative}. Specifically, given a trained generator \mbox{$p_\theta(x)=\int p_\theta(x|z)p(z)dz$} and a real valued discriminator function $d(x)$ (where higher $d(x)$ values indicate that the generated sample $x$ is closer to real data), an energy-based model can be defined as 
\mbox{$p_{\text{in}}(x)=\exp(d(x))p_\theta(x)/Z$}, where the use of $\exp(\cdot)$ %is to make sure 
ensures positive density. %is positive. 
In this case, the generator is also referred to as the `base distribution'~\cite{arbel2020generalized}, which estimates the support of the data distribution and the weight function $\exp(d(x))$ (corresponding to our semantic score function $s(x)$) %in our model) 
helps to `refine' the density on the support. %Different from 
In contrast to a GAN-based energy model, where the support of the base distribution $p_{\text{base}}$ is learned to be the support of the target data distribution $p_d$, the base distribution $p_{\text{base}}(x)$ in our model can alternatively cover a class of image distributions. For example, denote two image distributions as $p(x)$ and $q(x)$, our corresponding models are  defined as 
\begin{align}
    p(x) = \frac{p_{\text{base}}(x) s_p(x)}{Z_p}, \quad q(x) = \frac{p_{\text{base}}(x) s_q(x)}{Z_q}, 
\end{align}
where each model is assumed to lie on the support of $p_{\text{base}}(x)$, yet the semantic score functions $s_p(\cdot)$ and $s_q(\cdot)$ are specific to the respective distributions, see Figure \ref{eq:in:distribution} for a visualization.

%A lot of 
Several density-ratio based OOD methods can be unified under Equation~\ref{eq:in:distribution} and the corresponding score can be explained as the (unnormalised) density of a semantic distribution that is defined by Equation~\ref{eq:semantic:model}. In the following methods that we discuss, %below,
$p_{\text{in}}$ constitutes a generative model that is learned to fit the in-distribution data. Additionally various $p_{\text{base}}(x)$ have been proposed in the literature, which we also summarise below.

One definition of the base distribution $p_{\text{base}}(x)$ involves assigning \emph{positive density for images with valid local features}. In~\cite{zhang2021out}, a \emph{local model}, %which is 
designed to only be capable of capturing %to be only able to capture 
local pixel dependency, is learned to realise $p_{\text{base}}(x)$. Similarly, \cite{schirrmeister2020understanding,serra2019input} propose to use %a 
classic lossless compressors, \eg~PNG or FLIF, to play the role of the local model. Since the PNG or FLIF format only use the neighbouring pixels to predict the target pixel~\cite{boutell1997png}, the resulting coding length for a given data $x$ is approximately equal to the negative log-likelihood of a local model\footnote{See \cite{mackay2003information} for an introduction to the relationship between probabilistic modelling and lossless compression.}. In~\cite{havtorn2021hierarchical}, the base model is defined as a hierarchical VAE which ignores the latent variable that incorporates the high-level features, so that the positive mass is assigned to images with valid low-level (local) features. 

The base distribution $p_{\text{base}}(x)$ can also be defined to simply \emph{assign mass to all valid images in a certain domain}. For example, if the in-distribution data were to consist of images containing horses, the domain can be defined as the distribution of animals. In practice, $p_{\text{base}}(x)$ is learned to fit a large image dataset which can represent the domain. %For example, 
The work of~\cite{schirrmeister2020understanding} used Flow+~\cite{kingma2018glow} and PixelCNN~\cite{van2016pixel} models, %that are 
fitted to %a 
very large datasets, \eg~80 Million Tiny Images dataset~\citep{deng2009imagenet}. %to be the $p_{\text{base}}(x)$, 
We refer to such distributions as %the
\emph{universal models}.

% \begin{itemize}
%     % \item \textbf{Background as base:} The original paper that introduces likelihood ratio method \cite{ren2019likelihood} background as base 
%     \item \textbf{Local as base:} This assumes the base distribution gives positive mass to images that have valid local features. 
    
%     \cite{havtorn2021hierarchical,zhang2021out,schirrmeister2020understanding}
%     \item \textbf{All images as base:} (need to add this)
% \end{itemize}

In comparison with the approaches surveyed so far, we note that the likelihood ratio method proposed by~\cite{ren2019likelihood} does not fall under this framework. %They 
The authors alternatively assume that each data sample $x$ can be factorised into two distinct components \mbox{$x=\{x_b,x_s\}$}, where $x_b$ is a background component and $x_s$ a semantic component. Two independent models $p(x_b), p(x_s)$ are then trained to model the two respective components. In contrast, our framework assumes that both functions $p_{\text{base}}(x)$, $s(x)$ are supported on the $x$ space.

%\mt{Any other literatures using the likelihood ratio method?}

%It can be noted that 
We highlight that all discussed density estimation methods require the training of either one or two generative models. We argue that if the goal is to estimate the density ratio, training of complex generative models is not necessary. We propose to estimate the density ratio using the well-known class ratio estimation~\cite{sugiyama2012density,qin1998inferences,gutmann2010noise}, which only requires learning of a binary classifier and thus significantly simplifies the OOD detection workflow during both training and testing procedures.
In the following sections, we firstly give a brief introduction to class ratio estimation and then discuss practical considerations and applicability to % application and  for 
the OOD detection task.

\section{Model-Free Class Ratio Estimation}
\label{sec:ratioestimation}

We %first 
denote distributions $p_{\text{in}}(x)$ and $p_{\text{base}}(x)$ as two conditional distributions \mbox{$p(x|y=1)$} and \mbox{$p(x|y=0)$} respectively, such that the density ratio becomes
\begin{align}
   \frac{p_{\text{in}}(x)}{p_{\text{base}}(x)}=\frac{p(x|y=1)}{p(x|y=0)}.\label{eq:class:ratio}
\end{align}
We define a mixture distribution $p(x)$ as 
$p(x)\equiv p(x|y=1)p(y=1)+p(x|y=0)p(y=0)$,
where the Bernoulli prior distribution $p(y)$ represents the mixture proportions. We can  further assume a uniform prior $p(y=1)=p(y=0)=0.5$ and rewrite Equation \ref{eq:class:ratio} using Bayes rule
\begin{align}
    \frac{p_{\text{in}}(x)}{p_{\text{base}}(x)}&=\frac{p(x|y=1)}{p(x|y=0)}=\frac{p(y=1|x)\cancel{p(x)}}{\cancel{p(y=1)}}\Big/\frac{p(y=0|x)\cancel{p(x)}}{\cancel{p(y=0)}}=\frac{p(y=1|x)}{p(y=0|x)}.
    \label{eq:ratio}
\end{align}
We are then ready to estimate the ratio using a binary classifier. We initially sample labelled data from \mbox{$p(x,y)=p(x|y)p(y)$} by firstly sampling label $y'\sim p(y)$ and the corresponding data samples $x'\sim p(x|y=y')$. This is equivalent to sampling $x'\sim p_{\text{in}}$ when $y'=1$ and $x'\sim p_{\text{base}}$ when $y'=0$. The specified uniform prior $p(y)$ represents the probabilities to sample from $p_{\text{in}}$ or $p_{\text{base}}$, which are equal. The generated data pairs are then used to train a probabilistic classifier $p_\theta(y|x)$ with the cross entropy loss, which  has been shown to minimize the  Bregman divergence between the ratio estimation \mbox{$p_\theta(y=1|x)/ p_\theta(y=0|x)$} and the true density ratio~\cite{menon2016linking}.

After training the classifier, the density ratio estimator
\mbox{$\frac{p_{\theta}(y=1|x)}{1-p_\theta(y=1|x)}$}
can be used to perform OOD detection, thus avoiding the training of high-dimensional  generative models. It may be observed that the class ratio estimation scheme requires % the 
samples from both distributions; $p_{\text{in}}$ and $p_{\text{base}}$. The data samples of the in-distribution $p_{\text{in}}$ are already provided. We next discuss how to obtain samples from $p_{\text{base}}$, according to the differing base distribution assumptions that were discussed in Section~\ref{sec:ebm}.

\subsection{Construction of the base distribution}
\label{sec:ratioestimation:base}
As %we 
previously discussed, %in the previous section, 
training a binary classifier to estimate the ratio requires the samples from both $p_{\text{in}}$ and $p_{\text{base}}$. Samples from $p_{\text{in}}$ are just the in-distribution training dataset $\mathcal{D}_{\text{in}}$, which is given in the OOD detection task. We further discuss how to obtain the samples from $p_{\text{base}}$.

Recall, we %have discussed 
propose that %the 
existing OOD ratio methods can be viewed as building energy-based model with different $p_{\text{base}}$ distributions (Section~\ref{sec:ebm}). Specifically, these methods fall into two categories namely; (1) local model and (2) universal model %as the 
base distributions. Therefore, we propose two corresponding methods to construct the samples from $p_{\text{base}}$, to form a dataset $\mathcal{D}_{\text{base}}$.

\begin{itemize}
\item \textbf{Local  Model as Base Distribution}
In principle, one can train a local generative model~\citep{zhang2021out} on an image dataset and use that to generate local image samples, but this requires training a generative model. Alternatively, we propose to crop and resize the images from the  given in-distribution dataset $\mathcal{D}_{\text{in}}$. Intuitively, croping and resizing will preserve the local features, so the resulting images can be treated as the samples from a local model. We denote the resulting dataset as $\mathcal{D}_{\text{base}}^{\text{local}}$.
    \item \textbf{Universal Model as Base Distribution}
The construction of samples from the universal model is more straightforward. We can simply use a large image dataset, \eg~80 million tiny ImageNet ~\cite{hendrycks2018deep} as our base distribution. We denote this large image data by $\mathcal{D}_{\text{base}}^{\text{uni}}$.
\end{itemize}

%\looseness=-1
%Additionally, 
Under our  model assumption described in Section~\ref{sec:ebm}, the support of  $p_{\text{base}}$ should contain the support of $p_{\text{in}}$, we thus intentionally include the samples from $\mathcal{D}_{\text{in}}$ into the base distribution dataset by defining $\mathcal{D}_{\text{base}}=\mathcal{D}_{\text{base}}^{\text{local}}\cup \mathcal{D}_{\text{in}}$ or $\mathcal{D}_{\text{base}}=\mathcal{D}_{\text{base}}^{\text{uni}}\cup \mathcal{D}_{\text{in}}$. Further experimental details can be found in Section~\ref{sec:experiments}.

\subsection{Spread Density Ratio Score}
\label{sec:ratioestimation:semanticdensity}
The semantic score that is used for OOD detection is defined by the density ratio \mbox{$s(x)\propto p_{\text{in}}(x)/p_{\text{base}}(x)$}.
For a test data \mbox{$x_{\text{test}}\notin \textrm{supp}(p_{\text{base}})$}\footnote{We use $\textrm{supp}(p)$ to denote the support of distribution $p$.}, then \mbox{$p_{\text{base}}(x_{\text{test}})=0$} and the ratio is not defined. Ideally, we want $p_{\text{base}}(x_{\text{test}})$ to have support that covers all possible $x_{\text{test}}$. One solution is to add convolutional Gaussian noise $\tilde{p}_{\text{base}}=p_{\text{base}}*p_{\text{n}}$, 
where $p_n$ is an isotropic Gaussian distribution with mean 0 and variance $\sigma^2I_D$, with data space dimension $D$. %where $D$ is the dimension of the data space. 
However, when using class-ratio estimation, there is a danger that the classifier can easily distinguish between samples from two distributions by simply considering the noise level, resulting in a poor estimation of the density ratio. This %problem 
%is %also 
phenomenon is referred to as the ``density-chasm'' problem~\cite{rhodes2020telescoping}, in the class ratio estimation literature. To alleviate this problem, we propose to add the same convolutional noise to the distribution $p_{\text{in}}$: $\tilde{p}_{\text{in}}=p_{\text{in}}*p_n$. We can then define the \emph{spread density ratio score} $\tilde{s}$:
\begin{align}
   \tilde{s}(x)=\frac{\tilde{p}_{\text{in}}(x)}{\tilde{p}_{\text{base}}(x)}=\frac{(p_{\text{in}}*p_n)(x)}{(p_{\text{base}}*p_n)(x)}.
\end{align}
When $\sigma^2$ is small, we  assume that adding small pixel-wise noise to an image will not change the underlying semantics.  Therefore $\tilde{s}(x)$ can still provide a valid representation of the semantic score. The name \emph{spread density ratio} is inspired by recent work on {spread divergences}~\cite{zhang2020spread}, where convolutional noise is added to two distributions with different supports in order to define a valid KL divergence. Adding noise to the samples from two distributions has also been used to stabilize the training of GANs~\cite{sonderby2016amortised}.

When estimating the spread ratio $\tilde{s}(x)$ using the class-ratio estimation, we can sample from $\tilde{p}_{\text{in}}$ by first taking $x\sim p_{\text{in}}$, $\epsilon\sim p_n$ and let $\tilde{x}=x+\epsilon$ define a sample from $\tilde{p}_{\text{in}}$. The same scheme is used to construct the samples from $\tilde{p}_{\text{base}}$. Section 5.3 provides %further 
empirical evidence %showing 
to support the idea that %using 
the spread density ratio can significantly improve %the 
OOD detection results.

\section{Experiments}
\label{sec:experiments}
\begin{figure}[t]
\centering
  \includegraphics{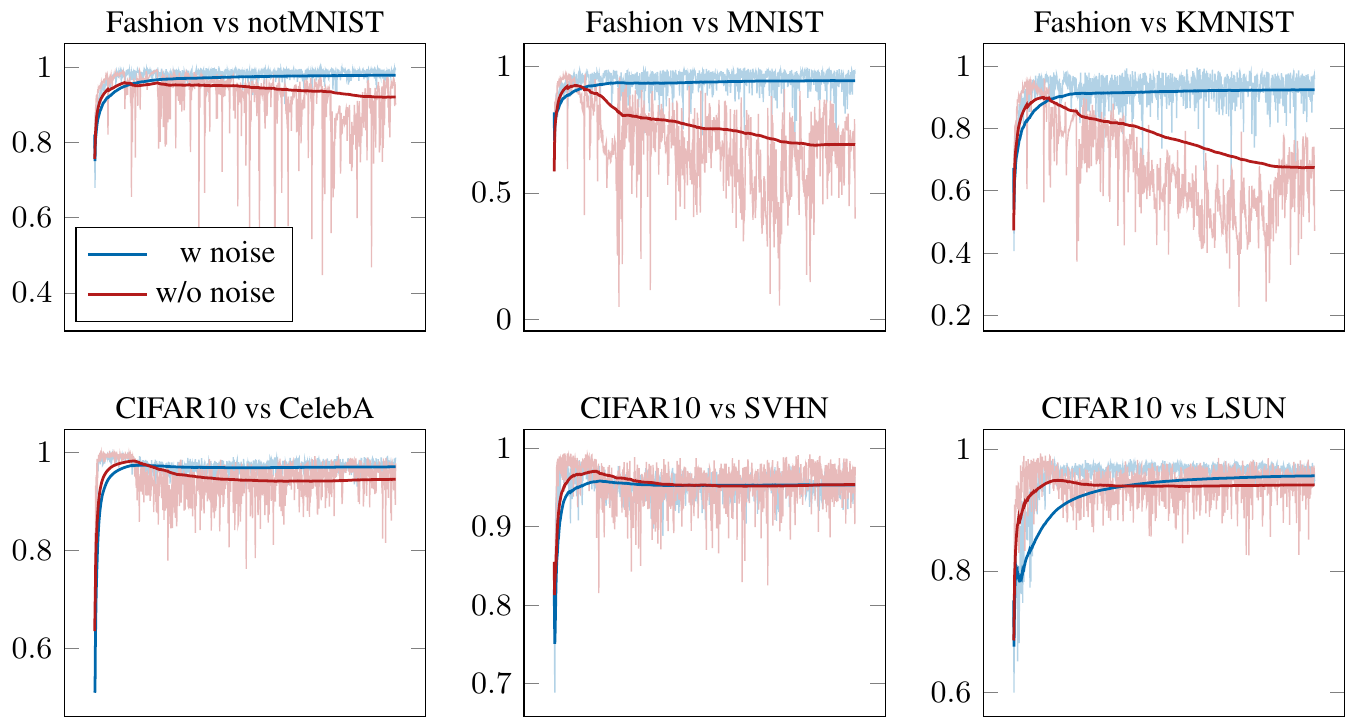} % faster
  \vspace{-1.7mm}
  \caption{AUROC per epoch. The $x$-axis represents 1000 training epochs and $y$-axis represents the AUROC. The model is trained on $\mathcal{D}_{\text{in}}$ = FashionMNIST / CIFAR10 and $\mathcal{D}_{\text{base}}$ = 80 Million Tiny Images, then tested on the corresponding OOD datasets. We show that adding convolutional noise results in significantly more stable AUROC results.\label{fig:noise}}
\captionof{table}{Comparison between local model and universal model as base distribution. From the table we can see that the universal model has better performance (AUROCs in \%) in most of the ID-OOD pairs. %The 
Results are averaged over five runs.}
  \label{table:ourresult}
  \centering
  \begin{minipage}{0.45\linewidth}
  \begin{small}
  \begin{tabular}{cccc}
    \toprule
    ID dataset & OOD& Local & Universal\\
    %\cmidrule(r){1-2}
    \midrule     
    \multirow{4}{*}{FMNIST}     &  MNIST &  73.1  &  \textbf{97.3} \\
     & NotMNIST & 89.2 & \textbf{99.3}  \\
    & KMNIST     & 69.0 & \textbf{95.8}   \\
    & Omniglot   & \textbf{100.0} & \textbf{100.0}  \\
    \midrule
    \multirow{4}{*}{MNIST}    &  FMNIST & 99.5 &  \textbf{100.0} \\
    & NotMNIST & 99.0 & \textbf{99.3} \\
    & KMNIST     & 90.1 & \textbf{95.8}   \\
    & Omniglot   & \textbf{100.0} & \textbf{100.0}  \\
    \bottomrule
    \end{tabular}
    \end{small}
  \end{minipage}
  \begin{minipage}{0.45\linewidth}
  \begin{small}
  \begin{tabular}{cccc}
    \toprule
    ID dataset & OOD& Local & Universal\\
    %\cmidrule(r){1-2}
    \midrule     
    \multirow{4}{*}{CIFAR10}    &  SVHN & 98.0 &  \textbf{98.2} \\
    & CIFAR100 & 54.7 & \textbf{85.9} \\
    & LSUN     & 37.0 & \textbf{97.3}   \\
    & CelebA   & 58.3 & \textbf{96.5}  \\
    \midrule
    \multirow{4}{*}{CIFAR100}    &  SVHN & \textbf{98.2} &  87.9 \\
    & CIFAR10 & 47.7 & \textbf{64.4} \\
    & LSUN     & \textbf{95.0} & 83.8   \\
    & CelebA   & 38.5 & \textbf{90.5}  \\
    \bottomrule
  \end{tabular}
  \end{small}
  \end{minipage}

\end{figure}
In this section, we %introduce the 
provide details of the implementations, the choice of datasets and the experimental results. The goal of our experiments is to demonstrate that for OOD detection tasks involving density ratios, a
sole use of a binary classifier can achieve competitive performance, in relation to % compared with %in comparison with the 
methods based on recent deep generative models. %based methods.

% \mt{need to re-construct this section a little bit:
% \begin{itemize}
%     \item network structure and training details
%     \item effectiveness of the spread ratio
%     \item our OOD detection results which includes the comparison between different $p_{\text{base}}$
%     \item compare with others
% \end{itemize}
% }
\begin{figure}[t]
\centering
\begin{tikzpicture}
\begin{groupplot}[group style={group size=2 by 2}, height=5cm, width=7.75cm, area style, ymajorticks=false, legend pos=north west, title style = {yshift=-0.65em}, legend cell align={left}]

\nextgroupplot[title={\small In-distribution: MNIST}]%, legend style={at={(0.5, 0.96)}, anchor=north}]
        \addplot+[ybar interval,mark=no, fill opacity=0.15, color=cornellred] plot coordinates { \input{data/hists/MNIST_test_scores.hist} };
        \addlegendentry{\tiny MNIST}
        \addplot+[ybar interval,mark=no, fill opacity=0.15, color=cornellblue] plot coordinates { 
\input{data/hists/MNIST_m_or_f_scores.hist} 
        };
        \addlegendentry{\tiny Fashion}
        \addplot+[ybar interval,mark=no, fill opacity=0.15, color=cornellgreen] plot coordinates { 
\input{data/hists/MNIST_notmnist_scores.hist} 
        };
        \addlegendentry{\tiny notMNIST}
        \addplot+[ybar interval,mark=no, fill opacity=0.15, color=mypurple] plot coordinates { 
\input{data/hists/MNIST_omni_scores.hist} 
        };
        \addlegendentry{\tiny Omniglot}
        \addplot+[ybar interval,mark=no, fill opacity=0.15, color=cornellgrey] plot coordinates { 
\input{data/hists/MNIST_kmnist_scores.hist} 
        };
        \addlegendentry{\tiny KMNIST}
\nextgroupplot[title={\small In-distribution: FashionMNIST}]%, legend style={at={(0.5, 0.96)}, anchor=north}]
        \addplot+[ybar interval,mark=no, fill opacity=0.15, color=cornellred] plot coordinates { \input{data/hists/FashionMNIST_test_scores.hist} };
        \addlegendentry{\tiny Fashion}
        \addplot+[ybar interval,mark=no, fill opacity=0.15, color=cornellblue] plot coordinates { 
\input{data/hists/FashionMNIST_m_or_f_scores.hist} 
        };
        \addlegendentry{\tiny MNIST}
        \addplot+[ybar interval,mark=no, fill opacity=0.15, color=cornellgreen] plot coordinates { 
\input{data/hists/FashionMNIST_notmnist_scores.hist} 
        };
        \addlegendentry{\tiny notMNIST}
        \addplot+[ybar interval,mark=no, fill opacity=0.15, color=mypurple] plot coordinates { 
\input{data/hists/FashionMNIST_omni_scores.hist} 
        };
        \addlegendentry{\tiny Omniglot}
        \addplot+[ybar interval,mark=no, fill opacity=0.15, color=cornellgrey] plot coordinates { 
\input{data/hists/FashionMNIST_kmnist_scores.hist} 
        };
        \addlegendentry{\tiny KMNIST}

\nextgroupplot[title={\small In-distribution: CIFAR10}]
        %\begin{axis}[
        %    width=5cm, height=5.3cm, area style,  legend pos=north east, font=\small, ymajorticks=false, legend cell align={left}]
        \addplot+[ybar interval,mark=no, fill opacity=0.15, color=cornellred] plot coordinates { \input{data/hists/cifar10_test_scores.hist} };
        \addlegendentry{\tiny CIFAR10}
        \addplot+[ybar interval,mark=no, fill opacity=0.15, color=cornellblue] plot coordinates { 
\input{data/hists/cifar10_svhn_scores.hist} 
        };
        \addlegendentry{\tiny SVHN}
        \addplot+[ybar interval,mark=no, fill opacity=0.15, color=cornellgreen] plot coordinates { 
\input{data/hists/cifar10_celeba_scores.hist} 
        };
        \addlegendentry{\tiny CelebA}
        \addplot+[ybar interval,mark=no, fill opacity=0.15, color=mypurple] plot coordinates { 
\input{data/hists/cifar10_lsun_scores.hist} 
        };
        \addlegendentry{\tiny LSUN}
        \addplot+[ybar interval,mark=no, fill opacity=0.15, color=cornellgrey] plot coordinates { 
\input{data/hists/cifar10_cifar100_scores.hist} 
        };
        \addlegendentry{\tiny CIFAR100}
        %\end{axis}
\nextgroupplot[title={\small In-distribution: CIFAR100}]
        \addplot+[ybar interval,mark=no, fill opacity=0.15, color=cornellred] plot coordinates { \input{data/hists/cifar100_test_scores.hist} };
        \addlegendentry{\tiny CIFAR100}
        \addplot+[ybar interval,mark=no, fill opacity=0.15, color=cornellblue] plot coordinates { 
\input{data/hists/cifar10_svhn_scores.hist} 
        };
        \addlegendentry{\tiny SVHN}
        \addplot+[ybar interval,mark=no, fill opacity=0.15, color=cornellgreen] plot coordinates { 
\input{data/hists/cifar100_celeba_scores.hist} 
        };
        \addlegendentry{\tiny CelebA}
        \addplot+[ybar interval,mark=no, fill opacity=0.15, color=mypurple] plot coordinates { 
\input{data/hists/cifar100_lsun_scores.hist} 
        };
        \addlegendentry{\tiny LSUN}
        \addplot+[ybar interval,mark=no, fill opacity=0.15, color=cornellgrey] plot coordinates { 
\input{data/hists/cifar100_cifar10_scores.hist} 
        };
        \addlegendentry{\tiny CIFAR10}
\end{groupplot}
\end{tikzpicture}
\vspace{-1.7mm}
\caption{Histograms of the log density ratios on the test datasets when $\mathcal{D}_{\text{base}}$ is a universal model. We use red to indicate the in-distribution test set $\mathcal{D}_{\text{in}}^{\text{test}}$ and other colours represent different OOD test datasets $\mathcal{D}_{\text{out}}^{\text{test}}$. The $x$-axis is the log density ratio and the $y$-axis is the correspond count. We observe, for relatively better cases (MNIST / FashionMNIST / CIFAR10 as in-distribution), the density ratios of $\mathcal{D}_{\text{in}}^{\text{test}}$ are concentrated, while in the CIFAR100 case, the density ratios of $\mathcal{D}_{\text{in}}^{\text{test}}$ are spread out, with a large overlap with the density ratios of $\mathcal{D}_{\text{out}}^{\text{test}}$, leading to a relatively small AUROC. 
  \label{fig:final hists}}
  
\captionof{table}{AUROC comparisons of approaches that use the additional 80 Million Tiny Images dataset, all results are  averaged over five runs.  %In this table, 
Both Outlier Exposure (OE)~\citep{hendrycks2018deep} and Tiny-Glow/PCNN~\citep{schirrmeister2020understanding} require training complex generative models.  We %can find 
observe our method achieves %can achieve 
competitive performance without requiring any generative model training.}%training any generative models.}
  \label{table:colorresult}
  \centering
  \begin{tabular}{cccccc}
    \toprule
    ID & OOD & OE \citep{hendrycks2018deep} & Tiny-Glow \citep{schirrmeister2020understanding} & Tiny-PCNN \citep{schirrmeister2020understanding}  & Ours \\
    %\cmidrule(r){1-2}
    \midrule     
    \multirow{3}{*}{CIFAR10}     &  SVHN     &  75.8 & 93.9 & 94.4  & \textbf{98.2} \\
    & CIFAR100 & 68.5 & 66.8 & 63.5 & \textbf{85.9} \\
    & LSUN     & 90.9 & 89.2 & 92.9 & \textbf{97.3} \\
    \midrule
    \multirow{3}{*}{CIFAR100}    &  SVHN     & - & 87.4 & \textbf{90.0}  & 87.9 \\
    & CIFAR10 & - & 52.8 & 54.5 &   \textbf{64.4} \\
    & LSUN     & - & 81.0 & \textbf{87.6} &  83.8 \\
    \bottomrule
  \end{tabular}

\end{figure}
\subsection{Neural Network Structure and Training Details}
\label{sec:experiments:outlier}

As introduced in Section~\ref{sec:ratioestimation}, our model is a binary classifier estimating \mbox{$p(y=1|x)$}. We use ResNet-18~\citep{he2016deep} for greyscale experiments and WideResNet-28-10~\citep{zagoruyko2016wide} for colour image experiments. The classifiers are trained for $1000$ epochs using a learning rate of $0.01$, batch size of $256$, and a Stocastic Gradient Descent(SGD) optimizer~\citep{robbins1951stochastic} with momentum %parameter 
= $0.9$.  The implementation can be found in our anonymous public repo\footnote{\url{https://github.com/andiac/OODRatio}}. All experiments are conducted on a NVIDIA Tesla V100 GPU.

Following recent work~\citep{hendrycks2016baseline}, we select MNIST~\citep{lecun1998gradient}, FashionMNIST~\citep{xiao2017fashion}, CIFAR10 and CIFAR100~\citep{krizhevsky2009learning} to be  in-distribution datasets $\mathcal{D}_{\text{in}}$, respectively. We use OMNIGLOT~\cite{lake2015human},  KMNIST~\citep{clanuwat2018deep} and notMNIST as OOD test datasets  $\mathcal{D}_{\text{out}}^{\text{test}}$ for greyscale experiments and we select SVHN~\cite{netzer2011reading}, LSUN~\citep{yu15lsun} and CelebA~\citep{liu2015faceattributes} as $\mathcal{D}_{\text{out}}^{\text{test}}$ for colour image experiments.

\subsection{Effectiveness of Spread Density Ratio}
\label{sec:experiments:spread}
%\looseness=-1
As discussed in Section~\ref{sec:ratioestimation:semanticdensity}, we add Gaussian noise to both $p_{\text{in}}$ and $p_{\text{base}}$ in the training stage and use the resulting spread density ratio to represent the semantic density. We apply Gaussian noise with standard deviation $0.1$ for both %of the 
greyscale and colour image experiments. For $p_{\text{base}}$, we use a universal model %for $p_{\text{base}}$ 
(described in Section~\ref{sec:ratioestimation:base}). %The details of how to construct samples from a 
Universal model sample construction details can be found in Section~\ref{sec:experiments:comparison}. %5.3.
Fig.~\ref{fig:noise} compares the test AUROC after each training epoch. We %can 
see that adding spread noise can significantly improve the distinguishability and training stability, for both datasets.

\subsection{Comparison Between Two Base Distributions}
\label{sec:experiments:comparison}

In this section, we compare two base distributions introduced in Section~\ref{sec:ratioestimation:base}: namely the local model and universal model. For the local model, samples are constructed by random cropping and resizing. In greyscale experiments, we crop the $28\times28$ images from $\mathcal{D}_{\text{in}}^{\text{train}}$ into $14\times14$ / $16\times16$ / \dots\,/ $24\times24$ / $26\times26$ images randomly, and then resize back to $28\times28$ using bilinear interpolation. Similarly, in colour image experiments, we crop the $32\times 32$ images into $16\times16$ / \dots\,/ $30\times30$ images randomly, then resize back to the original size, analogously. %by bilinear interpolation. 
The resulting images are denoted as $\mathcal{D}_{\text{base}}^{\text{local}}$. 

\begin{wrapfigure}{r}{0.5\textwidth}
\vspace{-0.5cm}
  \captionof{table}{AUROC comparisons. %with other approaches. 
  We %also 
  report the number of generative models used by alternative approaches in the third column. It may be observed that %We can see that 
  our model achieves relatively strong performance, (uniquely) without use of any generative models. Results for the Typicality test~\cite{serra2019input} %are from, 
  correspond to batches of two samples of the same type. %The 
  All results are averaged over five runs.} %of running five times.}
  \label{table:result}
  \begin{center}
  \begin{small}
  \vspace{-0,2cm}
  \begin{tabular}{lccc}
  \toprule
ID: & FMNIST & CIFAR10 & \multirow{1}{*}{Gen.} \\
  OOD : & MNIST  & SVHN \\
  \midrule
  WAIC \cite{choi2018waic}    & 76.6 &     \textbf{100.0} & 5 \\
  Like. Regret \citep{xiao2020likelihood} & 98.8 &  87.5  & 1 \\
  HVAE \citep{havtorn2021hierarchical} & 98.4 &  89.1 & 1 \\
MSMA KD \cite{mahmood2020multiscale} & 69.3 & 99.1 & 1\\

  OE \cite{hendrycks2018deep}    &  -  & 75.8 & 1 \\
  \hline
  \multicolumn{4}{c}{Density Ratio-based Methods}\\
  \hline
  Like. Ratio\cite{ren2019likelihood} & 99.7 &   91.2 & 2 \\
    Glow/PNG \cite{schirrmeister2020understanding} & -  &   75.4 & 1\\
  PCNN/PNG \cite{schirrmeister2020understanding} & -  &   82.3 & 1 \\
    Glow/FLIF \cite{serra2019input} & 99.8 &  95.0 & 1 \\
  PCNN/FLIF \cite{serra2019input} & 96.7  & 92.9 & 1 \\
  Global/Local\citep{zhang2021out}  & \textbf{100.0}  & 96.9 &2 \\
  Glow/Tiny \cite{schirrmeister2020understanding} & -   & 93.9 & 2 \\
  PCNN/Tiny \cite{schirrmeister2020understanding} & -   & 94.4 & 2  \\
  \textbf{Ours-Local}  &  73.1   & 97.2 & 0 \\
  \textbf{Ours-Universal} & 97.3 &  98.2  & 0 \\
  \bottomrule
  \end{tabular}
    \end{small}
  \end{center}
  \vspace{-1.cm}
\end{wrapfigure}
For the universal model, we use a cleaned\footnote{Cleaning follows the strategy suggested by~\citep{hendrycks2018deep} and removes images from the 80 Million Tiny Images dataset that belong to classes pertaining to CIFAR, LSUN and Places datasets.} subset of the 80 Million Tiny Images dataset~\citep{torralba200880}, resulting in $300K$ images~\citep{hendrycks2018deep}, that serve as our samples from a universal model, which ensures the diversity of
$\mathcal{D}_{\text{base}}^{\text{uni}}$~\citep{hendrycks2018deep}. For grey-scale experiments (\ie~when $\mathcal{D}_{\text{in}}$ is FashionMNIST or MNIST), we convert the 80 Million Tiny Images dataset to greyscale as our $\mathcal{D}_{\text{base}}^{\text{uni}}$.

Table~\ref{table:ourresult} shows the AUROC  comparison for two $p_{\text{base}}$ constructions. %where  we find that, the 
We find that our local model sample construction  %achieves good 
can achieve strong results in several cases; %such as 
CIFAR \emph{v.s.}~SVHN, MNIST \emph{v.s.}~all and FashionMNIST \emph{v.s.}~NotMNIST / Omniglot.  However, performance is more modest %it also performs poorly 
in several cases such as CIFAR10 \emph{v.s.}~CIFAR100 / LSUN / CelebA or CIFAR100 \emph{v.s.}~CIFAR10 / CelebA. 
In contrast, the samples from the universal model achieves strong performance in all experiments. We conjecture that our %local 
constructed local samples %still 
cannot comprehensively characterise the underlying local model (\ie~a model which can assign positive density to \emph{all} images with valid features). We believe the question of how to construct better local samples to be a promising future research direction.

\subsection{Comparisons with Other Methods}
We compare our approach, that defines $p_{\text{base}}$ using the universal model, %as  
with methods that also assume %have 
access to the Tiny-imagenet dataset, see Table~\ref{table:colorresult}. We observe that our method achieves relatively improved performance in four out of six ID-OOD pairs (see Fig.~\ref{fig:final hists} for corresponding histogram plots). %It is notable that, although 
Tiny-PCNN~\cite{schirrmeister2020understanding} achieves better performance in two data pairs, however, in contrast to our approach, we note that it requires training of two deep generative models.

We also compare our method to other recently proposed unsupervised OOD detection approaches, which %also includes the 
include density ratio methods. In  Table~\ref{table:result}, we %also 
report the number of generative models that each require to train. %are required to be trained, we can see that  
We observe that our method, with universal model, achieves competitive performance without training any generative models, providing computational efficiency. %thus resulting in a computationally efficient approach. %is computational efficient.

\section{Conclusion}
\label{sec:conclusion}
In this work, we propose an energy-based model framework that affords a unified modelling view of the recently proposed density-ratio based OOD methods. 
% Under our framework, the density ratios can be viewed as an unnormalized density from semantic models.
We further propose to use the class ratio estimation to estimate the density ratio, which does not require the training of complex generative models and yet can still achieve competitive OOD detection results, in comparison with state-of-the-art approaches. Our work gives rise to new potential directions \eg~more rigorous investigation of how to construct $p_{\text{base}}$, which we leave to future work.

\bibliographystyle{abbrvnat}
\bibliography{ref}

\newpage

\end{document}